\documentclass[10pt,twocolumn,letterpaper]{article}

\usepackage{cvpr}
\usepackage{times}
\usepackage{epsfig}
\usepackage{graphicx}
\usepackage{amsmath}
\usepackage{amssymb}

\usepackage{tabularx}
\usepackage{makecell}
\usepackage{multirow}
\usepackage{subfig}
\usepackage{tabulary}
\usepackage{booktabs}

\usepackage[breaklinks=true,bookmarks=false]{hyperref}

\cvprfinalcopy 

\def\cvprPaperID{****} 

\begin{document}

\title{PFDet: 2nd Place Solution to Open Images Challenge 2018 \\ Object Detection Track}
\author{
Takuya Akiba\thanks{The authors contributed equally and they are ordered alphabetically.} \,\,\,\,\,\,
Tommi Kerola$^*$ \,\,\,\, Yusuke Niitani$^*$ \,\,\,\, Toru Ogawa$^*$ \,\,\,\, Shotaro Sano$^*$ \,\,\,\, Shuji Suzuki$^*$ \\
Preferred Networks, Inc.\\
{\tt\small \{akiba,tommi,niitani,ogawa,sano,ssuzuki\}@preferred.jp}
}

\maketitle

\begin{abstract}
   We present a large-scale object detection system by team PFDet.
   Our system enables training with huge datasets using 512 GPUs,
   handles sparsely verified classes, and massive class imbalance.
   Using our method, we achieved 2nd place in the
   Google AI Open Images Object Detection Track 2018 on Kaggle.~\footnote{\url{https://www.kaggle.com/c/google-ai-open-images-object-detection-track}}
\end{abstract}

\vspace{-0.5cm}

\section{Introduction}
Open Images Detection Dataset V4 (OID)~\cite{openimages} is currently the largest publicly available object detection dataset, including $1.7$M annotated images with $12$M bounding boxes.
The diversity of images in training datasets is the driving force of the generalizability of machine learning models. Successfully trained models on OID would push the frontier of object detectors with the help of data.

Training a deep learning model on OID with low parallelization would lead to prohibitively long training times, as is the case for training with other large-scale datasets \cite{akiba2017extremely}.
We follow the work of MegDet~\cite{megdet} and use multi-node batch normalization to stably train an object detector with batch size of $512$.
Using ChainerMN~\cite{akiba2017chainermn}, a distributed deep learning library, we demonstrate highly scalable parallelization over $512$ GPUs.

OID is different from its predecessors, such as MS COCO~\cite{yi2014coco}, not merely in terms of the sheer number of images, but also regarding the annotation style.
In the predecessors, instances of all classes covered by the dataset are always exhaustively annotated, whereas in OID, for each image, instances of classes not verified to exist in the image are not annotated.
This is a realistic approach to expanding the number of classes covered by the dataset, because without sparsifying the annotated classes, the number of annotations required may explode as the total number of classes increases.

The problem with sparsifying the annotated classes is that most of the CNN-based object detectors learn by assuming that all regions outside of the ground truth boxes belong to the background.
Thus, in OID, these learning methods would falsely treat a bounding box as the background when an unverified instance is inside the box.
We find that the sparse annotation often leads to invalid labels, especially for classes that are parts of the other classes, which we call \emph{part classes} and \emph{subject classes}, respectively.
For instance, a human arm usually appears inside the bounding box of a person.
Based on this finding, we propose \textit{co-occurrence loss}.
For bounding box proposals that are spatially close to the ground truth boxes with a subject class annotation,  co-occurrence loss ignores all learning signals for classifying the part classes of the subject class.
This reduces noise in the training signal, and we found this leads to a significant performance improvement for part classes.

In addition to the previously mentioned uniqueness of OID, the dataset poses an unprecedented class imbalance for an object detection dataset.
The instances of the rarest class \textit{Pressure Cooker} are annotated in only $13$ images, but the instances of the most common class \textit{Person} are annotated in more than $800$k images.
The ratio of the occurrence of the most common and the least common class is $183$ times larger than in MS COCO~\cite{yi2014coco}.
Typically, this class imbalance can be tackled by over-sampling images containing instances of rare classes.
However, this technique may suffer from degraded performance for common classes, as the number of images with these classes decreases within the same number of training epochs.

As a practical method to solve class imbalance, we train models exclusively on rare classes and ensemble them with the rest of the models.
We find this technique beneficial especially for the first $250$ rarest classes, sorted by their occurrence count.

Our final model integrates solutions to the three noteworthy challenges of the OID dataset: a large number of images, sparsely verified classes, and massive class imbalance.
We use Feature Pyramid Network (FPN)~\cite{lin2017feature} with SE-ResNeXt-101 and SENet-154~\cite{hu2017squeeze} as backbones trained with sigmoid loss and cosine annealing as a learning rate scheduler~\cite{loshchilov2016sgdr}.

To summarize our major contributions:
\begin{itemize}
\item \textbf{Training at Scale}: We present the feasibility to train object detectors on a batch size of $512$ 
using ChainerMN~\cite{akiba2017chainermn} and 512 GPUs.
\item \textbf{Co-occurrence Loss}: We present \emph{co-occurrence loss} to ignore instances that are falsely labeled as negative for classes that are unverified using class-wise relationships constructed in advance.
\item \textbf{Expert Models}: We present the effectiveness of using expert models, especially for classes that rarely appear in the dataset.
\end{itemize}

\section{Method}
In this section we present our object detection system that allows fast large-scale training with high accuracy.

\subsection{Basic Architecture}
We use a two-stage Faster R-CNN style object detection framework~\cite{ren2015faster}
and leverage an SE-ResNeXt or SENet~\cite{hu2017squeeze} model as the backbone feature extractor.
To increase the global context information in the extracted features, we add an FPN and a pyramid spatial pooling (PSP)~\cite{zhao2017pyramid} module to the backbone.
Additionally, we increase the context information in the head network by concatenating features from twice the area around each RoI to the head before the fully-connected layers~\cite{zhu2017couplenet}.
We increase the number of scales of features extracted by the feature extractor to five from four, which is used in the original work of FPN~\cite{lin2017feature}.
This modification allows the network to gather even greater global context information.

Batch normalization (BN) is used ubiquitously to speed up convergence of training~\cite{ioffe2015batch}.
We use multi-node batch normalization~\cite{megdet} to share batch statistics of images across computing nodes so that the number of images used for collecting statistics is sufficiently large to stably compute the statistics.
To maximize the effectiveness of BN, we add BN layers to the top-down path of FPN in addition to the BN layers included in the base feature extractor, and the head network.

We train the CNN by first expanding ground truth labels to include all ancestor classes using the semantic hierarchy prepared in OID.
We formulate the learning problem as in a multi-label setting, and we use a sigmoid cross entropy loss for each class separately~\cite{uijlings2018revisiting}.
In the case when the ground truth class is not a leaf of the semantic hierarchy, we do not compute sigmoid cross entropy loss for descendants of the ground truth classes.

We use non-maximum weighted (NMW)~\cite{zhou2017cad} suppression during test time to reduce duplicate detections.
It was found that this works better than standard non-maximum suppression (NMS). NMS was used in the RPN while training.

\subsection{Co-occurrence Loss}
\begin{figure}
\subfloat[An image with annotations for some human parts (\textit{Human face} and \textit{Human arm}).]{
	\label{subfig:verified}
	\includegraphics[width=0.2\textwidth]{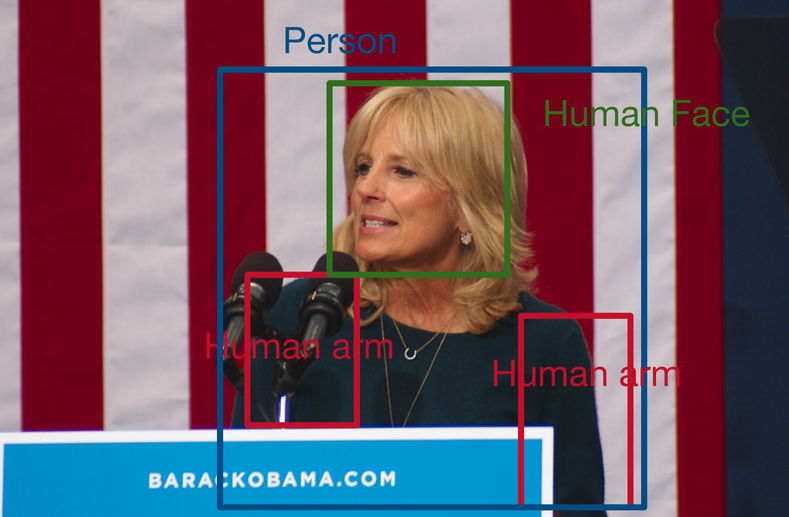} }
\hfill
\subfloat[An image with no verification of human part classes.]{
	\label{subfig:unverified}
	\includegraphics[width=0.2\textwidth]{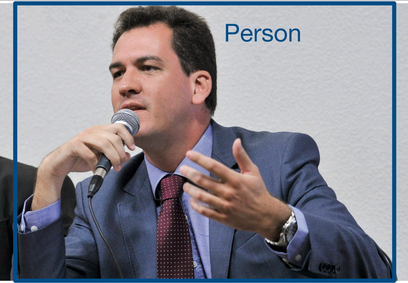} } 
\caption{Images and annotations from the OID dataset. In the right image (\ref{subfig:unverified}), even though \textit{Human face} clearly exists, there is no annotation because the class is not verified.}
\label{fig:sparse_verified}
\end{figure}

In the OID dataset, for each image, only the instances of classes that are verified are annotated.
In Figure~\ref{fig:sparse_verified}, images from OID with different coverage of verified labels are shown. In Figure~\ref{subfig:verified}, relatively more human part classes are labeled compared to Figure~\ref{subfig:unverified}, which has no annotation of human part classes such as \textit{Human face}.
With sparse annotations, conventional losses for training CNN-based object detectors~\cite{lin2017feature} would falsely label regions around instances as negatives.

To reduce false training signals, we introduce co-occurrence loss.
The main idea behind this loss is that for some classes, the relationship with other classes is informative enough that we can safely ignore false training samples.
For instance, given a ground truth bounding box of a person, it is highly likely that a human face exists inside the box even if human faces are not verified to exist in the image by the human annotator.
To implement the loss, we gather pairs of classes that satisfy the relationship ``if a proposal is inside a ground truth box of class $X$, it is safe to ignore treating the proposal as a negative sample of class $Y$''.
We use a pair when class $Y$ is a part of class $X$ or instances of class $Y$ is usually a possession of an instance of class $X$.
For instance, tires are parts of cars and jeans are usually possessed by a person.

\subsection{Expert Models}
In OID, there is an extreme class imbalance, which makes it difficult for a model to learn rare classes.
For instance, there are $238$ classes that are annotated in less than $1000$ images, but the most common class \textit{Person} is annotated in $807$k images.
We use expert models fine-tuned from a model trained with the entire dataset.
Each expert model is fine-tuned on a very small subset of the full category space, which we find to perform well for rare classes.

\subsection{Ensembling}
For the final submission, we use an ensemble of models trained on all $500$ classes and the expert models.
We do not apply duplicate suppression for individual models, but instead apply suppression once on the concatenation of the outputs of all models.

Since the distribution of the performance over classes is different among models, we prioritize outputs of models that are expected to perform better based on validation scores.
For each model $m$, we compute a weight $w_c^m$ for class $c$, which is multiplied to the confidence scores of the outputs of the model for this class.
Suppose that the mean of the validation scores of all models for class $c$ to be $\mu_c$, we set the weight $w_c^m$ to $\alpha$ if model $m$ performs lower than the average $\mu_c$.
Otherwise, we compute the weight by simply interpolating between $\alpha$ and $1$ linearly.
The computation is done as $w_c^m = \frac{s_c^m - \mu_c}{t_c - \mu_c} + \alpha \frac{t_c - s_c^m}{t_c - \mu_c}$,
where $s_c^m$ is the validation score of the model $m$ for class $c$ and $t_c$ is the highest validation score for class $c$.

\section{Experiments}
We used the split of the OID dataset for the ECCV2018 competition.
The recommended train and validation splits were used.
We never used the validation split to train networks.
The weights of the base feature extractor are pre-trained on ImageNet.
In addition to the OID dataset, we used MS COCO~\cite{yi2014coco} to train expert models for classes that are in the intersection of the label spaces of OID and MS COCO.



We use SGD with corrected momentum~\cite{goyal2017accurate} and a linear learning rate scaling rule with respect to the number of GPUs. The initial learning rate is set to $0.01$ for batch size of $8$.
The training starts with a warm-up phase.
Cosine annealing is used to attenuate the learning rate over time.

We scaled images during training so that the length of the smaller edge is between $[640, 1056]$. Also, we randomly flipped images horizontally to augment training data.
For the final submission, we augmented outputs at the test-time by concatenating outputs from inputs of multiple scales with and without horizontal flip.

\subsection{Software and Hardware Systems}
We use Chainer~\cite{tokui2015chainer} as our deep learning framework, ChainerMN~\cite{akiba2017chainermn} to scale up training and ChainerCV~\cite{niitani2017chainercv} for quick prototyping.
For training, we used MN-1b, an in-house cluster owned by Preferred Networks, Inc.
It consists of 64 nodes, where each node has two Intel Xeon Gold 6154 CPUs (3.0 GHz, 18 cores), 384 GB memory and eight NVIDIA Tesla V100 (32 GB memory). The nodes are interconnected by Mellanox Infiniband EDR.

\subsection{Results}
We first study the effectiveness of different techniques on the validation set.
Our baseline is FPN~\cite{lin2017feature} with SE-ResNeXt-101~\cite{hu2017squeeze} as the backbone trained with sigmoid loss for $12$ epochs.
The baseline model learning rate is multiplied by $\tfrac{1}{10}$ at epoch $8$ and $11$.
Mean average precision (mAP) of the validation split over $500$ classes is shown in Table~\ref{tab:ablative_single}.
For the final model, we increased the variation of anchors by using very tall and very wide ones.
The ratios of the width and height of the anchors are $[\tfrac{1}{3}, \tfrac{1}{2}, 1, 2, 3]$.
Using 512 GPUs, the computing time of training the final model was 33 hours for $16$ epochs. The scaling efficiency when using 512 GPUs was $83 \%$ in comparison to the single-node (i.e. 8 GPUs) baseline.

We also show the results after ensembling models in Table~\ref{tab:ablative_ensemble}.
Our final model outperformed the competition winner in the public leaderboard by $1.17$ mAP and falls behind them by only $0.03$ mAP on the private leader board.

In Table~\ref{tab:cooccur_class_score}, we show a comparison of a model trained with co-occurrence model and a model with identical setup except for the co-occurrence loss. For $47$ classes affected by co-occurrence loss, we see $9.2$ AP improvement on average.

In Table~\ref{tab:experts}, the ablative study of expert models is shown.
For the rarest 11-250 classes, we see improvement with expert models.
However, for more common classes (251-350), the fine-tuning has a negative effect.
Also, we consistently see better results if we reduce the size of the classes to which expert models are fine-tuned.

\begin{table}[t]
\centering \addtolength{\tabcolsep}{-2pt}
\footnotesize
\caption{Performance of a single model with single scale testing on the validation split.}\label{tab:ablative_single}
\begin{tabular}{l|c}
\toprule
 & validation mAP  \\
\midrule
Baseline (FPN with SE-ResNeXt-101) & 60.0  \\
+ multi-scale training & 60.3 (+0.3) \\
+ PSP and add BN to head & 60.4 (+0.1)  \\
+ Cosine Annealing & 63.4 (+3.0) \\
+ Add FPN scale & 64.5 (+1.1) \\
+ Co-occurrence loss & 65.2 (+0.7) \\
+ 16 epochs & 65.8 (+0.6) \\
+ Context head & 66.0 (+0.2) \\
+ SENet-154 and additional anchors & 67.5 (+1.5)  \\
\bottomrule
\end{tabular}\vspace{0.1cm}
\vspace{-0.03in}
\end{table}

\begin{table}[t]
\centering \addtolength{\tabcolsep}{-2pt}
\footnotesize
\caption{Ensemble of models with test-time augmentation.}\label{tab:ablative_ensemble}
\begin{tabular}{l|ccc}
\toprule
 & val mAP & Public LB & Private LB  \\
\midrule
Single best model & 69.95 & 55.81 & 53.43 \\
+ class20 experts & 71.73 & 59.34  & 55.87 \\
+ class10 experts & 72.33 & 60.19 & 56.61 \\
+ All the others except COCO & 73.98 & 61.83 & 57.97 \\
+ COCO & 74.07 & 62.34 & 58.48 \\
+ class-weight ensemble &  & 62.88 & 58.63 \\
\midrule
Competition winner &  & 61.71 & 58.66 \\
\bottomrule
\end{tabular}\vspace{0.1cm}
\vspace{-0.03in}
\end{table}

\begin{table*}[tp]
\centering
\caption{Ablative study of co-occurrence loss on classes that can be ignored by the loss. The scores are AP calculated on the validation set of the dataset.} \label{tab:cooccur_class_score}
\footnotesize \setlength{\tabcolsep}{2.3pt}         
\begin{tabulary}{\linewidth}{lcccccccccccccccc}\toprule[1pt]

& Arm & Ear & Nose & Mouth & Hair & Eye & Beard & Face & Head & Foot & Leg & Hand & Glove & Hat & Dress & Fedora  \\     \midrule

Baseline & $40.9$ & $17.5$ & $34.7$ & $21.4$ & $63.8$ & $27.3$ & $55.5$ & $82.7$ & $55.1$ & $50.7$ & $41.6$ & $32.3$ & $\mathbf{63.4}$ & $64.9$ & $70.6$ & $67.0$   \\
Co-occurrence & $\mathbf{55.2}$ & $\mathbf{62.6}$ & $\mathbf{69.6}$ & $\mathbf{55.2}$ & $\mathbf{74.7}$ & $\mathbf{64.0}$ & $\mathbf{76.8}$ & $\mathbf{91.4}$ & $\mathbf{78.9}$ & $\mathbf{59.5}$ & $\mathbf{54.4}$ & $\mathbf{53.6}$ & $60.8$ & $\mathbf{69.0}$ & $\mathbf{73.9}$ & $\mathbf{70.3}$   \\

\bottomrule
\toprule
& Footwe. & Sandal & Boot & Sports. & Coat & Sock & Glasse. & Belt & Helmet & Jeans & High h. & Scarf & Swimwe. & Earrin. & Bicycl. & Shorts \\     \midrule
Baseline & $61.9$ & $53.6$ & $\mathbf{61.6}$ & $52.9$ & $58.0$ & $\mathbf{70.6}$ & $74.9$ & $\mathbf{66.8}$ & $80.2$ & $62.7$ & $76.6$ & $71.6$ & $\mathbf{63.4}$ & $82.0$ & $75.1$ & $69.7$   \\
Co-occurrence & $\mathbf{68.5}$ & $\mathbf{58.9}$ & $57.9$ & $\mathbf{61.2}$ & $\mathbf{73.3}$ & $67.1$ & $\mathbf{85.4}$ & $61.9$ & $\mathbf{82.4}$ & $\mathbf{77.6}$ & $\mathbf{78.8}$ & $\mathbf{75.8}$ & $\mathbf{63.4}$ & $\mathbf{86.1}$ & $\mathbf{75.8}$ & $\mathbf{75.4}$  \\

\bottomrule
\toprule
& Baseba. & Minisk. & Cowboy. & Goggles & Jacket & Shirt & Sun ha. & Suit &  Trouse. & Brassi. & Tie & Licens. & Wheel & Tire & Handle & Average \\       \midrule 
Baseline & $\mathbf{67.2}$ & $\mathbf{62.5}$ & $65.0$ & $79.3$ & $69.5$ & $70.9$ & $61.3$ & $83.7$ & $62.5$ & $\mathbf{82.6}$ & $84.7$ & $72.1$ & $48.3$ & $49.4$ & $41.1$ & $61.1$   \\
Co-occurrence & $62.2$ & $58.7$ & $\mathbf{73.3}$ & $\mathbf{86.7}$ & $\mathbf{74.3}$ & $\mathbf{81.6}$ & $\mathbf{66.4}$ & $\mathbf{87.0}$ & $\mathbf{69.8}$ & $74.5$ & $\mathbf{91.5}$ & $\mathbf{74.6}$ & $\mathbf{66.4}$ & $\mathbf{69.6}$ & $\mathbf{46.2}$ & $\mathbf{70.3}$  \\
\bottomrule[1pt]
\end{tabulary} \vspace{-8pt} 
\end{table*} 

\begin{table}[t]
\centering \addtolength{\tabcolsep}{-2pt}
\footnotesize
\caption{Ablative study of expert models. Column "Index 11-100" lists the mean validation scores for the 11th class to the 100th class ordered by the occurrence in the dataset. Other columns similarly select the classes. The row "Full" lists scores of a model without fine-tuning. A row "ClassX experts" lists scores of expert models fine-tuned on class subsets of length X.}\label{tab:experts}
\begin{tabular}{l|ccc}
\toprule
 & Index 11-100 & Index 101-250 & Index 251-350  \\
\midrule
Full & 51.9 & 70.5 & \textbf{70.9} \\
Class10 experts & \textbf{65.6} & \textbf{73.1} & 66.3 \\
Class40 experts & 61.0 & 66.3 & 50.9 \\
\bottomrule
\end{tabular}\vspace{0.1cm}
\vspace{-0.03in}
\vspace{-0.25cm}
\end{table}

\section{Conclusion}
In this paper, we presented a large-scale object detector by team PFDet,
that allows scalable, fast object detection training on a large dataset
using 512 GPUs. The resulting fast research cycle allowed us to leverage
several techniques that led to 2nd place in the Google AI Open Images
Object Detection Track 2018 on Kaggle.
\vspace{-0.25cm}
{
\small
\paragraph{Acknowledgments}
We thank K.~Fukuda, K.~Uenishi, R.~Arai, S.~Omura, R.~Okuta, and T.~Abe for help with the experiments, and
R.~Calland for helping to improve the manuscript.
}

{\small
\bibliographystyle{ieee}
\bibliography{egbib}

\begin{thebibliography}{10}\itemsep=-1pt

\bibitem{akiba2017chainermn}
T.~Akiba, K.~Fukuda, and S.~Suzuki.
\newblock {ChainerMN: Scalable Distributed Deep Learning Framework}.
\newblock In {\em LearningSys workshop in NIPS}, 2017.

\bibitem{akiba2017extremely}
T.~Akiba, S.~Suzuki, and K.~Fukuda.
\newblock Extremely large minibatch {SGD:} {Training} {ResNet-50} on {ImageNet}
  in 15 minutes.
\newblock In {\em Deep Learning at Supercomputer Scale Workshop in NIPS}, 2017.

\bibitem{goyal2017accurate}
P.~Goyal, P.~Doll{\'a}r, R.~Girshick, P.~Noordhuis, L.~Wesolowski, A.~Kyrola,
  A.~Tulloch, Y.~Jia, and K.~He.
\newblock Accurate, large minibatch {SGD}: {Training} {ImageNet} in 1 hour.
\newblock {\em arXiv preprint arXiv:1706.02677}, 2017.

\bibitem{hu2017squeeze}
J.~Hu, L.~Shen, and G.~Sun.
\newblock Squeeze-and-excitation networks.
\newblock {\em CVPR}, 2018.

\bibitem{ioffe2015batch}
S.~Ioffe and C.~Szegedy.
\newblock Batch normalization: Accelerating deep network training by reducing
  internal covariate shift.
\newblock {\em ICML}, 2015.

\bibitem{openimages}
I.~Krasin, T.~Duerig, N.~Alldrin, V.~Ferrari, S.~Abu-El-Haija, A.~Kuznetsova,
  H.~Rom, J.~Uijlings, S.~Popov, S.~Kamali, M.~Malloci, J.~Pont-Tuset, A.~Veit,
  S.~Belongie, V.~Gomes, A.~Gupta, C.~Sun, G.~Chechik, D.~Cai, Z.~Feng,
  D.~Narayanan, and K.~Murphy.
\newblock Openimages: A public dataset for large-scale multi-label and
  multi-class image classification.
\newblock {\em Dataset available from
  https://storage.googleapis.com/openimages/web/index.html}, 2017.

\bibitem{lin2017feature}
T.-Y. Lin, P.~Doll{\'a}r, R.~B. Girshick, K.~He, B.~Hariharan, and S.~J.
  Belongie.
\newblock Feature pyramid networks for object detection.
\newblock In {\em CVPR}, 2017.

\bibitem{yi2014coco}
T.-Y. Lin, M.~Maire, S.~Belongie, L.~Bourdev, R.~Girshick, J.~Hays, P.~Perona,
  D.~Ramanan, C.~L. Zitnick, and P.~Dollár.
\newblock Microsoft coco: Common objects in context.
\newblock {\em ECCV}, 2014.

\bibitem{loshchilov2016sgdr}
I.~Loshchilov and F.~Hutter.
\newblock Sgdr: Stochastic gradient descent with warm restarts.
\newblock {\em ICLR}, 2017.

\bibitem{niitani2017chainercv}
Y.~Niitani, T.~Ogawa, S.~Saito, and M.~Saito.
\newblock Chainercv: a library for deep learning in computer vision.
\newblock In {\em ACM MM}, 2017.

\bibitem{megdet}
C.~Peng, T.~Xiao, Z.~Li, Y.~Jiang, X.~Zhang, K.~Jia, G.~Yu, and J.~Sun.
\newblock Megdet: A large mini-batch object detector.
\newblock In {\em CVPR}, 2018.

\bibitem{ren2015faster}
S.~Ren, K.~He, R.~Girshick, and J.~Sun.
\newblock Faster r-cnn: Towards real-time object detection with region proposal
  networks.
\newblock In {\em NIPS}, 2015.

\bibitem{tokui2015chainer}
S.~Tokui, K.~Oono, S.~Hido, and J.~Clayton.
\newblock Chainer: a next-generation open source framework for deep learning.
\newblock In {\em LearningSys workshop in NIPS}, 2015.

\bibitem{uijlings2018revisiting}
J.~Uijlings, S.~Popov, and V.~Ferrari.
\newblock Revisiting knowledge transfer for training object class detectors.
\newblock In {\em CVPR}, 2018.

\bibitem{zhao2017pyramid}
H.~Zhao, J.~Shi, X.~Qi, X.~Wang, and J.~Jia.
\newblock Pyramid scene parsing network.
\newblock In {\em CVPR}, 2017.

\bibitem{zhou2017cad}
H.~Zhou, Z.~Li, C.~Ning, and J.~Tang.
\newblock Cad: Scale invariant framework for real-time object detection.
\newblock In {\em ICCV Workshops}, 2017.

\bibitem{zhu2017couplenet}
Y.~Zhu, C.~Zhao, J.~Wang, X.~Zhao, Y.~Wu, H.~Lu, et~al.
\newblock Couplenet: Coupling global structure with local parts for object
  detection.
\newblock In {\em ICCV}, 2017.

\end{thebibliography}
}

\end{document}